# 3 Dimensional Dense Reconstruction: A Review of Algorithms and Dataset

Yangming Li

*Abstract*—3D dense reconstruction refers to the process of obtaining the complete shape and texture features of 3D objects from 2D planar images. 3D reconstruction is an important and extensively studied problem, but it is far from being solved. This work systematically introduces classical methods of 3D dense reconstruction based on geometric and optical models, as well as methods based on deep learning. It also introduces datasets for deep learning and the performance and advantages and disadvantages demonstrated by deep learning methods on these datasets.

## I. INTRODUCTION

3D dense reconstruction is a computer vision technique that involves the creation of a 3D model of an object or scene from a set of 2D images or video frames. The goal is to estimate the 3D geometry and appearance of the object or scene at every point in space, resulting in a detailed and dense 3D model.

3D dense reconstruction is a critical step in many applications. For example, in robotics, it is critical to tasks such as robot navigation, environment understanding, object manipulation and etc[1–6]. In mixed reality, 3D reconstruction is used to create virtual objects or environments, which are needed to be integrated with the real world[7–9]. In computer vision, 3D reconstruction is used to create 3D models for object recognition, tracking, and pose estimation[10, 11].

The study of 3D dense reconstruction has a long history, and its origins can be traced back to the development of photogrammetry in the mid-19th century[12]. Photogrammetry is the science of making measurements from photographs, and it is based on the principles of triangulation and stereoscopy[13]. These principles were used to create 3D models of terrain and buildings from aerial photographs, and the field of photogrammetry grew rapidly in the early 20th century. In the 1970s and 1980s, researchers began to develop techniques for reconstructing 3D shapes from multiple images. These early methods were limited to simple shapes and relied on hand-crafted features and heuristics[14]. With the advent of digital cameras and the increasing availability of powerful computing resources in the 1990s and 2000s, the field of 3D dense reconstruction began to grow rapidly[15]. Researchers developed a range of new techniques, including structure from motion[16], multi-view stereo[17], and dense depth estimation[18], which allowed for more accurate and detailed 3D reconstruction from images. Starting from the early 2010s, deep learning starts to evolve and advance 3D dense reconstruction, and researchers developed a range of new architectures and techniques to learn to predict depth or surface normals from input images, and have shown promising results in improving precision, increasing reliability, and expanding the applicability.

Datasets are crucial to these data-driven deep-learning models. Actually, one of the biggest challenges in 3D dense reconstruction with deep neural networks is obtaining high-quality training data. This requires large datasets with diverse scenes and objects, captured from a variety of viewpoints and under different lighting conditions. Ideally, the dataset should also be annotated with ground-truth 3D models, which can be used to train and evaluate the performance of the deep neural network. In recent years, several large-scale datasets have been created for 3D dense reconstruction with deep neural networks, such as the ShapeNet dataset, the ScanNet dataset, etc[19, 20].

This work reviews triangulation- and stereovision-based algorithms, deep-learning-based algorithms, and popular datasets to provide an up-to-date, accurate, and concise research status, help identify areas for improvement, and help establish best practices and standard field of study. This work explains why deep learning has dominated the field and examines the intrinsic connections between datasets and algorithms to reveal gaps in the types of scenes or objects represented, or inconsistencies in how data is labeled or annotated , and stimulated work leading to more accurate and reliable 3D reconstruction results.

This work is organized as following: the second section presents Geometrical Model based 3D Reconstruction, including Structure from Motion, Shape from Shade, Simultaneous Localization and Mapping, explains the problem-solving philosophy behind these methods with example algorithms. Section 3 presents Deep Learning based 3D Dense Reconstruction, including Convolutional Neural Networks, Recurrent Neural Networks, Graph Neural Networks, Generative Adversarial Networks (GANs), Autoencoders and Variational Autoencoders (VAEs), explains the architectural differences of these methods and discusses some representative algorithms in each category. Chapter 4 introduces some widely used datasets for deep 3D reconstruction, and summarizes the performance of some deep learning algorithms on popular dataset. The last chapter discusses the limitations of deep learning based 3D dense reconstruction, and presents potential solutions that can further revolutionize 3D dense reconstruction.

Electrical and Computer Engineering Technology Department, Rochester Institute of Technology, Rochester, NY, USA 14623

## II. GEOMETRICAL MODEL BASED 3D RECONSTRUCTION

### A. Overview

The 3D dense reconstruction process typically involves the following steps:

Image Acquisition: The first step is to capture multiple images or video frames of the scene or object from different angles and viewpoints. The quality and resolution of the input images significantly impact the accuracy of the final 3D model.

Feature Detection and Matching: In this step, distinctive features or keypoints are identified in the input images, and correspondences between these features in different images are established. Common feature detection algorithms include SIFT, SURF, and ORB.

Camera Pose Estimation: Once the feature correspondences are found, the relative position and orientation of the cameras are estimated, allowing for the reconstruction of the scene's geometry. This process is often performed using methods such as the essential matrix, homography matrix, or bundle adjustment.

Depth Estimation: Depth information is calculated for each pixel in the input images, often using stereo matching or multi-view stereo algorithms. This step creates a dense point cloud, which represents the 3D structure of the scene or object.

Surface Reconstruction: The dense point cloud is then converted into a 3D mesh, which represents the surface of the object or scene. This can be achieved using algorithms like Poisson Surface Reconstruction or Marching Cubes.

Texturing: Finally, the color and texture information from the input images are projected onto the 3D mesh, resulting in a photorealistic 3D model.

Overall, 3D dense reconstruction enables the creation of highly accurate and detailed 3D models from 2D images, providing valuable insights and immersive experiences across various industries and applications.

### B. Structure from Motion

Structure from Motion (SfM) is a computer vision technique that aims to estimate the 3D structure of a scene or object from a sequence of 2D images or video frames taken from different viewpoints[16]. It involves estimating the camera motion (poses) and the 3D coordinates of the scene points by analyzing the apparent motion of the object points in the image sequence (Fig. 1). SfM is widely used in applications like robotics, aerial mapping, and 3D reconstruction.

The typically SfM pipeline is nearly identical to the general 3D reconstruction pipeline: it first gathers images or video frames from various angles, then identify features using algorithms like SIFT, SURF, or ORB[21–23], and match keypoints across images using algorithms like nearest neighbor and joint compatibility tests[24–26]. Based on the feature associations, it determines camera positions and orientations using techniques like essential matrix or bundle adjustment, and triangulates features to create a sparse 3D point cloud. For the dense reconstruction, it interpolates sparse point cloud using methods like stereo matching, and converts point cloud to 3D mesh with algorithms like Poisson Surface Reconstruction[27]. Structure from Motion enables the reconstruction of 3D structures from 2D images while accounting for camera motion, making it a powerful tool for various applications across numerous industries.

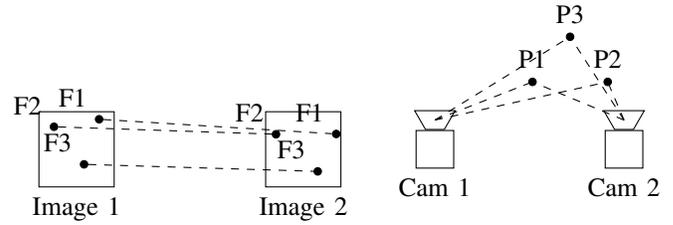

Fig. 1: Structure from Motion.

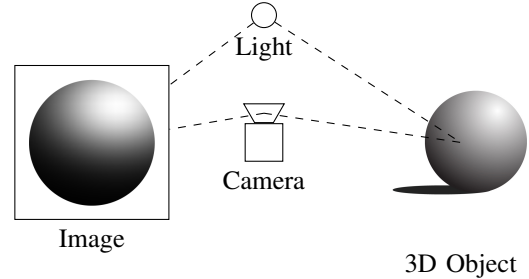

Fig. 2: Shape from Shade.

### C. Shape from Shade

Shape from Shading (SfS) recovers the 3D shape or geometry of an object or scene from a single 2D image by analyzing the shading or intensity variation in the image [28]. The shading information is influenced by the interaction between the surface geometry, illumination, and the material properties of the object (Fig. 2). By making certain assumptions about these factors, Shape from Shading aims to infer the depth information and reconstruct the 3D structure of the scene.

The assumptions usually made in SfS include:

- Lambertian Surface: The object's surface is assumed to have a Lambertian reflectance model, meaning it reflects light equally in all directions, and the intensity of the reflected light depends only on the angle between the light source and the surface normal[28].

- Known Lighting Conditions: The position, intensity, and color of the light source(s) are assumed to be known or estimated.

- Smooth Surface: The object's surface is assumed to be smooth, with continuous variations in depth and surface normals.

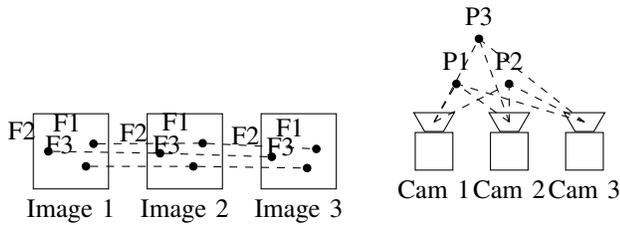

Fig. 3: Simultaneous Localization and Mapping for dense visual reconstrucgtion.

- Single Image: SfS works with a single image, unlike other 3D reconstruction techniques that rely on multiple images or stereo pairs.

Because SfS adopts a different technology with SfM, the SfS process is drastically different. It often first preprocess the input image reduce noise and enhance the shading information. Secondly, based on the shading information and the assumptions made, the surface normals at each pixel in the image are estimated. Thirdly, the estimated surface normals are integrated to recover the depth information, resulting in a 3D representation of the object or scene. This step can involve solving a partial differential equation (PDE), optimization, or other numerical techniques [28].

Shape from Shading has some limitations, such as sensitivity to noise, ambiguities in reconstructing the 3D shape, and reliance on the assumptions made. However, it remains an important technique for 3D reconstruction, especially in situations where only a single image is available, and has been used in applications like robotics, medical imaging, and object recognition.

*D. SLAM*

Simultaneous Localization and Mapping (SLAM) is a technique used in robotics and computer vision to estimate the camera's pose (localization) and create a 3D map of the environment (mapping) simultaneously[29]. SLAM-based 3D dense reconstruction aims to generate a detailed and accurate three-dimensional representation of the scene or object in real-time as the camera moves through the environment.

To a certain degree, SLAM can be viewed as an extension of Structure from Motion (SfM) technology to encompass global landmarks[30–32]. This is accomplished by way of the continuous identification and updating of global landmarks utilizing Bayesian networks or similar techniques, and through the loop closing, the error accumulation is maximally suppressed (Fig. 3). Some of typical SLAM-based dense reconstruction methods are described in the following.

KinectFusion is a real-time 3D dense reconstruction algorithm that uses depth data captured by a Microsoft Kinect sensor[33]. It generates a dense 3D model by continuously fusing depth measurements into a global 3D model, represented as a truncated signed distance function (TSDF) volume. The camera's pose is estimated by aligning the current depth frame with the global model using an iterative closest point (ICP) algorithm.

ElasticFusion is an approach that combines dense SLAM with non-rigid surface deformation[34]. It builds a global map represented as a surfel-based model while continuously estimating the camera's pose. The method can handle large-scale environments and loop closures, allowing for better consistency in the reconstructed 3D model.

Large-Scale Direct Monocular SLAM (LSD-SLAM) is a monocular SLAM method that operates directly on the image intensity values, avoiding the need for feature extraction and matching[35]. It estimates the camera's pose and a semi-dense depth map in real-time, creating a 3D model of the environment. LSD-SLAM can be extended to RGB-D data (color and depth) for more accurate dense reconstructions.

ORB-SLAM is a feature-based SLAM method that uses ORB (Oriented FAST and Rotated BRIEF) features for localization and mapping[23]. Although it primarily generates a sparse 3D map, it can be extended with dense reconstruction techniques like stereo matching or multi-view stereo to create dense 3D models.

Dense Tracking and Mapping (DTAM) is a monocular SLAM method that estimates dense depth maps for each frame in real-time by minimizing the photometric error between the input image and a rendered version of the 3D model[36]. The depth maps are then merged into a global 3D model, and the camera's pose is estimated using a dense direct tracking method.

These SLAM-based 3D dense reconstruction methods offer real-time performance and can be employed in various applications, such as robotics, autonomous navigation, augmented reality, and virtual reality. They can handle complex environments and camera motion, making them well-suited for generating 3D models in dynamic and real-world scenarios.

III. DEEP LEARNING BASED 3D DENSE RECONSTRUCTION

Deep learning techniques for 3D dense reconstruction emerged by the inspiration that the model-based algorithms show limited adaptiveness and robustness in real-world applications, and as a result of advancements in computer vision, graphics, and machine learning.

Currently, deep learning based 3D dense reconstruction does demonstrate their advantages of a higher accuracy, improved robustness, and scalability. These algorithms will be summarized below according to the types of techniques they adopted.

*A. Convolutional Neural Networks*

CNNs are widely used for image-based tasks due to their ability to capture local and global features in images, and their advantages of being scale-invariant and location-invariant[37–40]. They have been adapted for depth estimation, surface normal prediction, and semantic segmentation, which can be combined to perform dense 3D reconstruction.

Deep Multi-view Stereo (DeepMVS) is one of the representative works that use CNN for dense reconstruction[41]. DeepMVS takes a reference image and neighboring images with known camera parameters as input. The architecture comprises shared-weight CNNs for feature extraction, cost volume construction for similarity calculation, 3D CNN-based cost volume regularization, and depth map prediction using sub-pixel interpolation (Fig. 4). DeepMVS is trained end-to-end with a multi-scale loss function, achieving superior performance on various benchmarks.

DenseDepth uses a convolutional neural network to predict depth maps from RGB images, but from single image[42]. The model leverages both local and global features to improve depth estimation accuracy. Structure from Motion Learner (SfMLearner) is an unsupervised learning framework for depth and camera pose estimation from monocular videos[43]. The model consists of two CNNs: one for depth estimation and one for pose estimation. It uses the photometric consistency loss between the input image and a synthesized image to learn depth and ego-motion simultaneously. SfMLearner does not require any ground truth depth or pose annotations, which makes it suitable for large-scale training on real-world videos.

*B. 3D Convolutional Neural Networks (3D-CNNs)*

3D-CNNs are an extension of traditional CNNs that operate on volumetric data, making them suitable for processing 3D information. They can be used for voxel-based representations, point cloud processing, and 3D segmentation.

Volumetric U-Net (3D U-Net) adapts the popular U-Net architecture for dense 3D segmentation tasks [44]. The network has three components, the encoder, the decoder and the final layer (Fig. 5). The encoder consists of a series of 3D convolutional layers followed by 3D max-pooling layers. The number of feature maps is doubled after each max-pooling operation. The decoder consists of a series of 3D up-convolutional layers followed by concatenation with the corresponding feature maps from the encoder (skip connections), and 3D convolutional layers. The number of feature maps is halved after each up-convolutional operation. The final layer is a 1x1x1 3D convolutional layer that maps the feature maps to the desired number of output channels (i.e., segmentation classes).

Octree-based CNN (OctNet) utilizes an octree-based representation to enable efficient learning of 3D structures at high resolutions[45]. It is more memory-efficient compared to regular 3D CNNs, but its performance depends on the quality of the octree construction. It can handle sparse annotations and provide accurate reconstructions, but it might require more memory and computational resources compared to OctNet[44, 45]. Voxel-based 3D CNN (VoxelNet) transforms point clouds into a voxel representation and processes them using a 3D CNN for object detection tasks[46]. While it is effective in detecting objects within point clouds, it might not be as suitable for dense reconstruction tasks as the other two methods.

*C. Recurrent Neural Networks (RNNs) and Long Short-Term Memory (LSTM)*

RNNs and LSTMs are particularly useful for processing sequences of data, which is widely used in robotics, and can be beneficial in 3D dense reconstruction tasks with temporal components, such as video-based reconstruction or SLAM[47–52]. They can be used to model temporal dependencies and improve the accuracy of the reconstruction[53]. Existing methods use RNN to restore motion trajectories, so these methods usually also use other architectures, such as CNN, to achieve complete 3D dense reconstruction.

Deep Visual Odometry (DeepVO) estimates camera motion and depth from video sequences using CNNs and RNNs[51]. DeepVO contains two main components: feature extraction and pose estimation (Fig. 6). The feature extraction module consists of a Convolutional Neural Network (CNN) that takes a pair of consecutive monocular images as input and generates feature maps. The CNN is pretrained on the ImageNet dataset and fine-tuned for the visual odometry task. The pose estimation module is based on a Recurrent Neural Network (RNN) with Long Short-Term Memory (LSTM) units. It takes the feature maps generated by the feature extraction module as input and predicts the 6-DOF camera pose. The RNN is able to model temporal dependencies between consecutive frames, which is crucial for visual odometry tasks.

Depth and Motion Network (DeMoN) uses both RNN and other architectures to estimates depth and camera motion from a pair of images[54]. It consists of two main components: the Depth and Motion Network (DMN) and the Iterative Refinement Network (IRN). The DMN is composed of two convolutional neural networks (CNNs) that share the same architecture but have different weights. The IRN consists of several convolutional layers and a recurrent neural network (RNN) module. The RNN module, which is a Convolutional LSTM (ConvLSTM), is used to model the iterative refinement process. Compared to DeepMVS, DeMoN might be more accurate in estimating depth and motion compared to DeMoN, but is less generalizable[41, 54]. Unsupervised Deep Visual Odometry (UnDeepVO) adapts the SfMLearner's unsupervised approach to stereo camera setups, improving the depth estimation accuracy[55]. However, it requires stereo input data and might still be less accurate than supervised techniques.

*D. Graph Neural Networks (GNNs)*

GNNs are designed to operate on graph-structured data, making them suitable for processing irregular data structures like point clouds. They have been applied to tasks like point cloud segmentation, classification, and completion.

PointNet is a deep learning architecture for point cloud processing that uses shared multi-layer perceptrons to learn point-wise features and global max-pooling to aggregate features[56]. The primary goal of PointNet is to learn a global point cloud feature that is invariant to permutations, translations, and rotations. The architecture consists of the

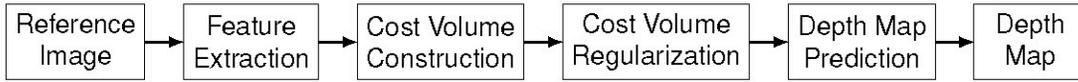
Fig. 4: Architecture of Deep Multi-view Stereo.

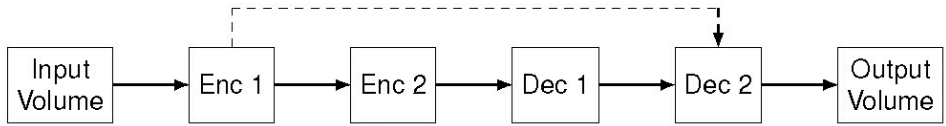
Fig. 5: Architecture of 3D UNet.

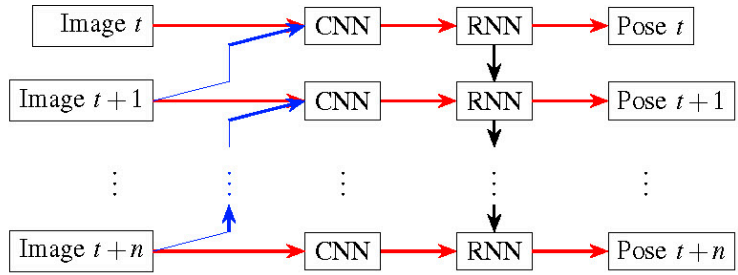
Fig. 6: Architecture of Deep Visual Odometry.

following 5 components (Fig. 7). 1, the input is an unordered set of 3D points, where each point is represented by its x, y, and z coordinates. 2, Transformation Networks (T-Nets) are mini-PointNets that learn spatial transformations to align the input point cloud. There are two T-Nets in the architecture: the first one predicts a 3x3 transformation matrix to align the point cloud, and the second one predicts a 64x64 transformation matrix to align the features. 3, Multi-Layer Perceptrons (MLPs) are fully connected layers that learn local features for each input point. The architecture contains several MLP layers, each with a varying number of neurons (e.g., 64, 128, or 1024). The MLPs apply a shared weight function to each point independently, making the architecture permutation invariant. 4, Max Pooling is the symmetric function is applied to aggregate the local features into a global point cloud feature. Max pooling is used as the symmetric function, which allows the network to capture the most salient features of the input point cloud. 5, Fully Connected Layers and Output (MLPs) processes the global point cloud feature to generate the final output. For classification tasks, the output layer has as many neurons as there are object classes, while for segmentation tasks, the output layer generates per-point scores.

Dynamic Graph CNN (DGCNN) constructs dynamic graphs based on nearest neighbors in the input data, allowing the network to adapt to varying point cloud densities[57]. It offers improved classification and segmentation performance compared to PointNet but can be computationally expensive due to graph construction. PointNet++ extends the original PointNet architecture by applying it hierarchically to learn local and global features[58]. It achieves better performance in dense reconstruction tasks compared to PointNet, but might be less flexible than DGCNN in handling varying point densities.

*E. Generative Adversarial Networks (GANs)*

GANs consist of a generator and a discriminator network that compete with each other, enabling the learning of complex data distributions. GANs have been applied to tasks like image synthesis, inpainting, and depth map refinement in dense 3D reconstruction.

3D Generative Adversarial Network (3D-GAN) generates 3D shapes by learning a mapping from random noise to 3D volumes[59]. The model comprises two primary components: a generator and a discriminator (Fig. 8). The generator is a 3D convolutional neural network (CNN) that takes a random noise vector as input and generates a 3D object as output. The generator upsamples the input noise vector using transposed 3D convolutional layers followed by Batch Normalization and ReLU activation functions. The architecture of the generator resembles a 3D U-Net, with skip connections between the corresponding layers. The discriminator is another 3D CNN that takes the generated 3D object as input and classifies it as either real (from the training dataset) or fake (generated by the generator). The architecture consists of several 3D convolutional layers followed by Batch Normalization, Leaky ReLU activation functions, and a final fully connected layer with a sigmoid activation function. It can produce diverse shapes but might lack fine details and accuracy in complex scenes.

Pixel-to-Voxel Reconstruction (Pix2Vox) transforms 2D images into 3D voxel representations using a 3D encoder-decoder network and a 2D encoder network[60]. It is designed for single-view reconstruction and can produce high-quality results, but it relies on voxel representations, which can be memory-intensive.

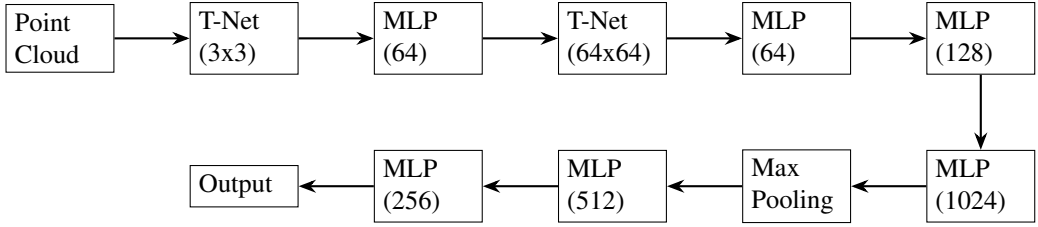
Fig. 7: Architecture of PointNet.

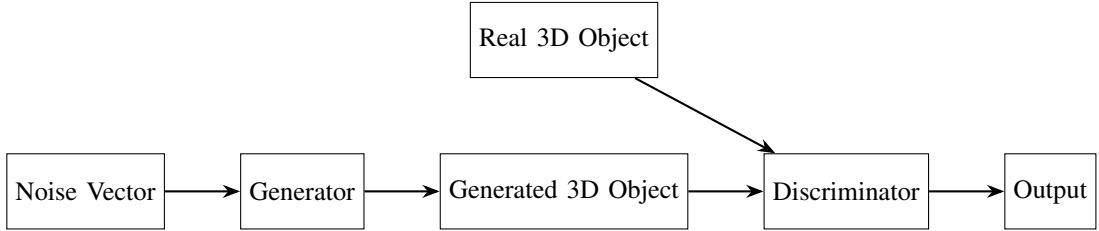
Fig. 8: Architecture of 3D Generative Adversarial Network.

*F. Autoencoders and Variational Autoencoders (VAEs)*

Autoencoders and VAEs are unsupervised learning methods that learn to encode and decode data, capturing the underlying structure and distribution of the data. They can be employed for tasks like feature learning, denoising, and reconstruction.

3D-Autoencoder learns a compact representation of 3D data (point clouds, meshes, or voxel grids) by encoding and decoding input data[61]. The model comprises two primary components: a generator and a discriminator, both of which are 3D convolutional neural networks (CNNs) (Fig. 9). The generator is a 3D CNN that takes a random noise vector as input and generates a 3D object as output. The generator upsamples the input noise vector using transposed 3D convolutional layers followed by Batch Normalization and ReLU activation functions. The architecture of the generator resembles a 3D U-Net, with skip connections between the corresponding layers. The discriminator is another 3D CNN that takes the generated 3D object as input and classifies it as either real (from the training dataset) or fake (generated by the generator). The architecture consists of several 3D convolutional layers followed by Batch Normalization, Leaky ReLU activation functions, and a final fully connected layer with a sigmoid activation function. It can produce accurate reconstructions but might be limited in its ability to generate novel shapes.

3D Variational Autoencoder extends the 3D autoencoder by adding a probabilistic layer, which enables the generation of new shapes by sampling from the latent space[62]. It offers more flexibility in generating novel shapes but might produce less accurate reconstructions compared to 3D autoencoders.

These deep neural network architectures can be used individually or combined in various ways to achieve the desired level of detail and accuracy in 3D dense reconstruction tasks, depending on the specific requirements and the data available.

However, the performance of the deep learning models relies heavily on the availability and quality of training data. Obtaining well-annotated data for 3D reconstruction tasks can be challenging. Besides, deep learning models are often considered "black boxes," making it difficult to understand their inner workings and troubleshoot potential issues.

## IV. DATASET FOR DEEP LEARNING BASED 3D DENSE RECONSTRUCTION

A large number of dataset are available for training and testing deep learning based 3D dense reconstruction algorithms (Table I). These dataset has its unique characteristics, making them suitable for different tasks and algorithms. Factors such as scene diversity, data quality, and availability of ground truth data should be considered when choosing a dataset for a specific 3D dense reconstruction task. For example, KITTI, SUN3D, and Matterport3D are suited for evaluating dense reconstruction, and Middlebury and ETH3D are used widely for 3D reconstruction but are also suitable for evaluating stereo and multi-view stereo algorithms. This work will discuss the characteristics and compare some of the algorithms on dataset for improving the accuracy and efficiency of researchers selecting datasets and benchmark algorithms.

*A. Dataset Review*

The ShapeNet dataset (https://shapenet.org/) is a large-scale collection of 3D models that was created in 2015, but keeps growing dataset[19]. The dataset consists of over 51,300 unique 3D models that are organized into various categories such as chairs, cars, airplanes, and more. Each model in the dataset is represented as a collection of 3D meshes that can be rendered from various viewpoints using a graphics engine to produce 2D images.

Middlebury Stereo Dataset (http://vision.middlebury.edu/stereo/) is a relatively small indoor dataset with ground truth depth maps obtained using a structured light scanner [63]. This

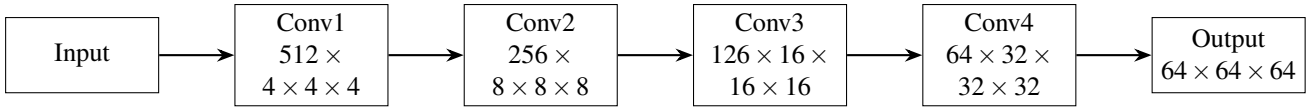

Fig. 9: Architecture of the encoder of 3D-Autoencoder. The decoder has the same architecture.

TABLE I: Dataset for 3D Dense Reconstruction

| Dataset | Year of creation | Size and Type of Scenes | Size | Source of Depth | Camera Pose |
|---|---|---|---|---|---|
| ShapeNet | 2015 | 31,350, In & Out | 300 M | Synthetic | No |
| Middlebury Stereo | 2001 | 47, In | 47 pairs | Structured light, Stereo | Yes |
| KITTI Vision | 2012 | ∼28, Out | 42,382 | Velodyne LiDAR | Yes |
| ETH3D | 2017 | 27, In & Out | 27 sets | Laser scanner | Yes |
| NYU Depth V2 | 2012 | 1,449, In | 144,959 | Kinect | Yes |
| SUN3D | 2013 | 415, In & Out | N/A | Kinect, Xtion | Yes |
| TUM RGB-D | 2012 | 39, In | N/A | Kinect | Yes |
| ICL-NUIM | 2014 | 8, In | N/A | Synthetic | Yes |
| EuRoC MAV | 2016 | 11, In | N/A | Laser scanner | Yes |
| ApolloScape | 2018 | N/A, Out | >140,000 | LiDAR | Yes |
| ScanNet | 2017 | 2,513, In | N/A | Kinect v2, RealSense | Yes |
| Matterport3D | 2017 | 90, In & Out | N/A | Matterport camera | Yes |
| Stanford 2D-3D-S | 2017 | 6 areas, In | 70,496 | Matterport camera | Yes |
| SceneNet RGB-D | 2016 | 5 million, In | 5 million | Synthetic | Yes |
| Sintel | 2010 | N/A, In & Out | 1,064 | Synthetic | No |
| Redwood | 2016 | 100, In | N/A | Structure sensor | Yes |
| FlyingThings3D | 2016 | N/A, In & Out | 3,720 | Synthetic | Yes |
| 7-Scenes | 2014 | 7, In | N/A | Kinect | Yes |
| Washington RGB-D | 2011 | 300, In | N/A | Kinect | Yes |
| Blensor | 2013 | N/A, In & Out | N/A | Synthetic | Yes |
| DTU Robot | 2014 | 124, In | 5,000+ | Structured light | Yes |
| Stanford 3D | 2006 | N/A, In & Out | N/A | Range scans | Yes |
| Freiburg Forest | 2016 | 1, Out | N/A | Stereo | Yes |
| SCARED | 2017 | 7, Med | 15,000 | Kinect/Synthetic | Yes |
| EndoSLAM | 2016 | 35, Med | 60,000 | CT | Yes |

Note: In, Out, and Med denotes Indoor, Outdoor, and Medical scenes, respectively.

dataset contains high-quality images and depth maps, which make it suitable for benchmarking stereo algorithms. The Middlebury Stereo Dataset is widely used for benchmarking stereo matching algorithms. The dataset provides an online leaderboard, ranking algorithms based on their performance. The leaderboard gets updated frequently as new algorithms are submitted.

ETH3D (https://www.eth3d.net/) contains both indoor and outdoor scenes with high-resolution images and ground truth depth maps[64]. This dataset contains challenging scenes with diverse environments and varying lighting conditions; useful for evaluating multi-view stereo and 3D reconstruction algorithms.

KITTI Vision Benchmark Suite (http://www.cvlibs.net/datasets/kitti/) is a large dataset with outdoor street scenes captured using a vehicle-mounted stereo camera rig[65]. It contains high-precision ground truth depth data, which is captured from LIDAR sensors, but is limited to outdoor street scenes.

SUN3D (http://sun3d.cs.princeton.edu/) contains RGB-D video sequences of indoor scenes captured using a Microsoft Kinect sensor[66]. It provides camera pose information and contains diverse set of scenes, but its depth data is generated by the Kinect and has lower precision compared to the depth measured by laser sensors.

Matterport3D (https://niessner.github.io/Matterport/) is a large-scale RGB-D dataset of indoor scenes with 10,800 panoramic views from 194 different spaces[67]. The dataset contains high-resolution images, depth maps, and camera poses, as well as diverse set of scenes with varying object types and room layouts.

DTU Robot Image Dataset (http://roboimagedata.compute.dtu.dk/) is a multi-view stereo dataset with 60 indoor scenes captured using a robotic arm-mounted camera[68]. The dataset contains high-resolution images and accurate camera poses, for indoor object-centered scenes.

BlendedMVS (https://github.com/YoYo000/BlendedMVS) is a large-scale multi-view stereo dataset with 17,000 high-resolution images of 1,000 diverse objects[69]. One of the uniqueness of this dataset is that it has mix of synthetic and real-world images, which focuses on object reconstruction rather than scene reconstruction.

Tanks and Temples (https://www.tanksandtemples.org/) is a dataset with both indoor and outdoor scenes featuring complex geometry and diverse objects[70]. The dataset contains both high-quality images and ground truth point clouds generated using a laser scanner.

ScanNet (http://www.scan-net.org/) is a large-scale RGB-D dataset with 2.5 million views in more than 1,500 indoor

scenes[20]. Both high-quality images, depth maps, and camera poses are contained for indoor scenes.

TUM RGB-D Dataset (https://vision.in.tum.de/data/datasets/rgbd-dataset) contains RGB-D video sequences captured using a Microsoft Kinect sensor and camera poses from a motion capture system in various indoor environments[71].

NYU Depth Dataset V2 (https://cs.nyu.edu/~silberman/datasets/nyu_depth_v2.html) contains both RGB-D data of captured using a Microsoft Kinect sensor in indoor environments, and labeled pairs of aligned RGB and depth images[72].

SceneNN (http://www.scenenn.net/) is a large dataset that contains annotated RGB-D images of indoor scenes, and groundtruth annotations for segmentation and 3D reconstructions[73].

Stanford 2D-3D-Semantics (2D-3D-S) Dataset (https://3dscenegraph.stanford.edu/) is a dataset containing RGB-D images, 3D point clouds, and semantic annotations for 6 large-scale indoor spaces. The dataset has diverse set of scenes with varying object types and room layouts.

MegaDepth Dataset (http://www.cs.cornell.edu/projects/megadepth/) is a large-scale dataset with diverse scenes, and contains Internet photos of various scenes, and reconstruction results using Structure-from-Motion and Multi-View Stereo[74].

ApolloScape (http://apolloscape.auto/) is a large-scale dataset for autonomous driving research, including semantic segmentation, instance segmentation, 3D car instance reconstruction, and visual localization[75]. The dataset contains both high-quality images and annotations, and has diverse set of scenes and objects.

OpenRooms (https://vilab-ucsd.github.io/ucsd-openrooms/) is a large-scale dataset with diverse room layouts, and contains synthetic indoor scenes with RGB-D images, semantic annotations, and camera poses[76].

Stanford Light Field Archive (http://lightfield.stanford.edu/lfs.html) contains light field images of various scenes and objects, and provides unique data for research in light field applications.

Freiburg Forest Dataset (http://deepscene.cs.uni-freiburg.de/) contains RGB, depth, and thermal images captured in a forest environment using a custom sensor setup[77].

3D60: Indoor Scene Understanding in 3D (https://vcl3d.github.io/3D60/) is a dataset containing 60 indoor scenes with 360° panoramas, depth maps, and semantic annotations[78]. The dataset has high-quality images and depth maps, suitable for evaluating 3D reconstruction, scene understanding, and navigation tasks, diverse set of scenes with varying object types and room layouts.

Taskonomy Dataset (http://taskonomy.stanford.edu/) is a large scale dataset that contains 4 million images from 20,000 scenes[79]. The dataset contains 3D point clouds, surface normals, and semantic annotations.

Replica Dataset (https://github.com/facebookresearch/Replica-Dataset) is a dataset of high-quality 3D reconstructions of indoor environments, including RGB-D images, semantic annotations, and camera poses[80].

SCARED dataset (The stereo correspondence and reconstruction of endoscopic data) (https://endovissub2019-scared.grand-challenge.org/) is a medical dataset captured using a da Vinci Xi surgical robot. The dataset consists of 7 training datasets and 2 test datasets, and each of the dataset corresponds to a single porcine subject, in which an endoscope was moved using the da Vinci Xi to observe scenes, and the camera's pose relative to the fixed robot base was recorded based on robot kinematic status[81].

EndoSLAM dataset (https://github.com/CapsuleEndoscope/EndoSLAM) is a medical dataset that consists of both ex-vivo and synthetically generated data. The ex-vivo part of the dataset includes standard as well as capsule endoscopy recordings. The dataset is divided into 35 sub-datasets. Specifically, 18, 5 and 12 sub-datasets exist for colon, small intestine and stomach respectively[82].

*B. Algorithms and Dataset*

The performance of deep learning-based dense reconstruction algorithms is heavily dependent on the quality and size of the dataset used for training. A large and diverse dataset with accurate ground truth depth information and camera pose information is crucial for training deep learning models for dense reconstruction.

We summarize the performance of some of the algorithms on NYU Depth V2 dataset (Table II), KITTI vision dataset (Table III), and TUM RGB-D dataset (Table IV) to demonstrate the performance of state of art techniques. Please notice that the summary does not mean to be a comparison on performance, instead, it means to show the magnitude of problems, because these algorithms focus on different tasks within 3D dense reconstruction, which make a direct comparison less meaningful.

V. DISCUSSION

Although deep learning-based algorithms have shown significant progresses and almost dominate the 3D dense reconstruction field in recent years, there are still several limitations to these approach.

Deep learning-based algorithms require large amounts of diverse training data, which can be time-consuming and expensive to obtain. Deep learning-based algorithms rely on high-quality training data to achieve high performance but it is nearly impossible to achieve the accurate ground-truth in depth-sensor denial and time-varying scenes, such as endoscopic surgeries[82].

Due to the limitations of existing techniques, there are some challenges that desire novel and powerful solutions.

| Algorithm | RMSE (m) | Rel Error | $\sigma < 1.25$ | $\sigma < 1.25^2$ | $\sigma < 1.25^3$ |
|---|---|---|---|---|---|
| [83] | 0.641 | 0.214 | 0.611 | 0.887 | 0.971 |
| [84] | 0.573 | 0.127 | 0.811 | 0.953 | 0.988 |
| [85] | 0.523 | 0.120 | 0.838 | 0.976 | 0.997 |
| [86] | - | - | 0.821 | 0.965 | 0.995 |
| [87] | 0.471 | 0.187 | 0.815 | 0.955 | 0.988 |
| [88] | - | - | 0.852 | 0.970 | 0.994 |

TABLE II: NYU Depth V2 dataset.

| Algorithm | RMSE (m) | Rel Error | $\sigma < 1.25$ | $\sigma < 1.25^2$ | $\sigma < 1.25^3$ |
|---|---|---|---|---|---|
| [89] | 6.266 | 0.203 | 0.696 | 0.900 | 0.967 |
| [90] | 4.627 | 0.117 | 0.845 | 0.951 | 0.984 |
| [91] | 4.863 | 0.187 | 0.809 | 0.953 | 0.986 |
| [92] | 4.459 | 0.115 | 0.861 | 0.961 | 0.986 |
| [93] | 4.401 | 0.112 | 0.868 | 0.967 | 0.991 |

TABLE III: KITTI vision dataset.

| Algorithm | Performance |
|---|---|
| [94] | RMSE: 0.573 |
| [95] | Acc.: 72.34% |
| [96] | ATE: 0.0177 |
| [97] | ATE: 0.0135 |
| [98] | ATE: 0.0189 |
| [99] | EPE: 0.0163 |
| [100] | ATE: 0.0165 |

TABLE IV: TUM RGB-D dataset. RMSE: root-mean-square error), Acc: accuracy, ATE: absolute trajectory error, and EPE: endpoint error.

Low texture: In scenes with little or no texture (e.g., flat, homogeneous surfaces), it becomes difficult to establish feature correspondences between different views. This can lead to incorrect depth estimation and incomplete or noisy reconstructions[101, 101–105].

Dynamic objects: Objects in the scene that move between frames or views introduce inconsistencies in the reconstruction process. As depth estimation and correspondence matching rely on the assumption that the scene is static, dynamic objects can cause errors in the reconstructed geometry[106–110].

Low image quality: Images with low resolution, noise, or poor lighting conditions can affect the performance of feature detection and matching algorithms, leading to incorrect depth estimation and inaccurate reconstructions. High-quality images are essential for robust 3D dense reconstruction[111–115].

Deformation: Non-rigid or deformable objects, such as fabric or humans, can cause inconsistencies in the reconstruction process. Deformation may change an object's appearance between views, making it challenging to establish correct feature correspondences and to estimate the object's true 3D structure[116–118].

Drastic scene depth changes: Scenes with a significant variation in depth, such as indoor environments with close-up objects and faraway walls, can cause difficulties in depth estimation and feature matching. Algorithms need to adapt to these variations to produce accurate reconstructions[103, 119, 120].

Motion blur: Fast-moving objects or camera motion can cause motion blur in the images, making it difficult to detect and match features accurately. This can lead to incorrect depth estimation and reconstruction artifacts[103, 121–123].

Adverse illumination conditions: Challenging lighting conditions, such as shadows, glare, or over- or under-exposure, can negatively impact feature detection and matching algorithms. Reflective or transparent surfaces may cause misleading feature matches due to changes in appearance depending on the viewpoint. This can lead to incorrect depth estimation and reconstruction artifacts[124–126]. Robust algorithms need to handle these conditions to ensure accurate reconstruction.

We summarize some of the existing solutions that address one of more of these problem in Table V.

We believe that there are several potential solutions that can further improve the robustness, precision, and reliability of 3D dense reconstruction in real-world applications.

Multi-modal data fusion: Combining data from different sensors such as cameras, LiDAR, and IMU can improve the accuracy and reliability of the 3D reconstruction. This approach can also provide more robustness against sensor failures or limitations.

Incremental reconstruction: Incremental reconstruction techniques can improve the efficiency and robustness of 3D reconstruction in dynamic scenes. Instead of processing the entire scene at once, the reconstruction is performed incrementally, updating the reconstruction as new data becomes available.

Incorporating prior knowledge: Incorporating prior knowledge about the scene or the object being reconstructed can improve the accuracy and robustness of 3D dense reconstruction. This can include using known object shapes, known camera trajectories, or incorporating physical constraints such as object rigidity.

Hybrid methods: Hybrid methods that combine deep learning-based algorithms with traditional computer vision techniques can improve the accuracy and robustness of 3D dense reconstruction. For example, combining deep learning-based feature extraction with traditional structure-from-motion techniques can improve the accuracy of camera pose estimation.

|  | Static | Dynamic object | Low texture | Image quality | Illumination | Recovery | Motion | Deformation | Scene depth |
|---|---|---|---|---|---|---|---|---|---|
| sparse | [23, 127] | [109, 128–130] | [131, 132] | [133] | [34, 132] | [134, 135] | [136] | | |
| semidense | [137, 138] | [139] | [140] | [137] | | | | | |
| full-dense | [141] | [106–108] | [101–104] | [111, 113] | [124–126] | | [103] | [116, 117] | [103, 119] |

TABLE V: Example algorithms that address challenges in real-world 3D reconstruction.

Active learning: Active learning techniques can help to reduce the amount of labeled data required for training the deep learning model. This can reduce data acquisition costs and improve the scalability of the approach.

Particularly, we believe causal deep learning, a relatively new approach that emphasizes the importance of causality in modeling complex systems. Recently Deep Causal Learning has demonstrated its capability of using prior knowledge to disentangle modeling problems and reduce data needs, improve performance on extrapolating unseen data, modularize learning problems, and incrementally learn from multiple studies, therefore, can fundamentally overcome the data scarcity problem and improve 3D dense reconstruction. [142–144].